\documentclass[a4paper]{article}                                    

\usepackage{a4wide}

\usepackage{hyperref}       
\usepackage{url}            
\usepackage{booktabs}       
\usepackage{amsfonts}       
\usepackage{nicefrac}       
\usepackage{microtype}      
\usepackage{graphicx}
\usepackage{amssymb}
\usepackage{amsfonts}
\usepackage[cmex10]{amsmath}

\DeclareMathOperator*{\argmax}{\arg\!\max}

\title{Self-supervised learning: When is fusion of the primary and secondary sensor cue useful?}

\author{
  G.C.H.E. de Croon
}

\begin{document}

\maketitle

\begin{abstract}
Self-supervised learning (SSL) is a reliable learning mechanism in which a robot enhances its perceptual capabilities. Typically, in SSL a trusted, primary sensor cue provides supervised training data to a secondary sensor cue. In this article, a theoretical analysis is performed on the fusion of the primary and secondary cue in a minimal model of SSL. A proof is provided that determines the specific conditions under which it is favorable to perform fusion. In short, it is favorable when (i) the prior on the target value is strong or (ii) the secondary cue is sufficiently accurate. The theoretical findings are validated with computational experiments. Subsequently, a real-world case study is performed to investigate if fusion in SSL is also beneficial when assumptions of the minimal model are not met. In particular, a flying robot learns to map pressure measurements to sonar height measurements and then fuses the two, resulting in better height estimation. Fusion is also beneficial in the opposite case, when pressure is the primary cue. The analysis and results are encouraging to study SSL fusion also for other robots and sensors.
\end{abstract}

\section{Introduction}

As robots move more and more into dynamic real-world environments, learning mechanisms are getting increasingly important. However, learning robots are held back by multiple issues, including: potential unreliability of the learning process, long learning times, and the requirement of intensive human supervision. In light of these issues, an interesting learning mechanism is \emph{Self-Supervised Learning} (SSL). It focuses on the augmentation of a robot's perception capabilities. Typically, in SSL the robot uses a trusted, primary sensor cue to train a secondary sensor cue with supervised learning. If learning is successful, the secondary sensor will give similar outputs to the primary sensor cue. For example, the car that won the grand DARPA challenge, Stanley \cite{THRUN2006} used a laser scanner as the primary sensor to classify areas ahead as being part of the road or not. It used a color camera as the secondary sensor, and learned a mapping from colors to the class labels ``road'' or ``not road''. Since the camera could evaluate the terrain much further into the distance than the (range-limited) laser scanner, Stanley could drive faster. This was important for winning the race.

SSL has the following beneficial properties: (i) the robot always keeps access to the trusted primary cue, which can be used to ensure the safety of the system during and after learning, (ii) learning is supervised, which means that it is relatively fast and can build on an enormous amount of research in machine learning, (iii) since the supervised targets are provided by a robotic sensor, no human supervision is required and an ample amount of training data is available for the machine learning algorithms such as deep neural networks (e.g., \cite{lecun1995convolutional,krizhevsky2012imagenet}).

The main reason to perform SSL is that the two sensor cues have complementary properties. The example of Stanley showed that the secondary cue may have a longer range than the primary cue (see also \cite{lieb2005adaptive,lookingbill2007reverse,hadsell2009learning,muller2013real,lamers2016self}). In the literature, other types of complementarities have been studied as well. For instance, in \cite{baleia2015exploiting} a robot first judges terrain traversability by means of haptic interaction, and uses SSL to learn this same capability with the camera. Another interesting example is given in \cite{ho2015optical}, in which a flying robot selects a landing site by making use of optical flow from an onboard camera as the primary cue. The secondary sensor cue consists of image appearance features, which after learning allows the robot to select a landing site without moving.



\begin{figure}[t!]
\centering
\includegraphics[width=8cm]{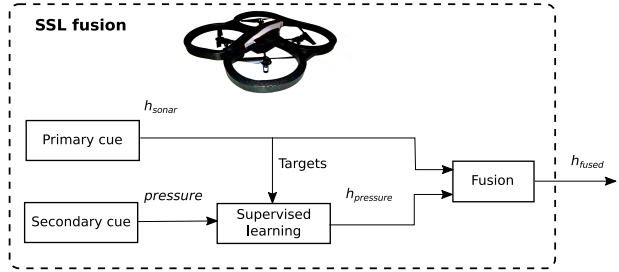}
\caption{Illustration of fusion in self-supervised learning, using the case study from this article. The robot uses a trusted primary sensor cue (sonar height measurements in this case) to train a secondary sensor cue (pressure measurements). After training, the robot has learned to transform the raw pressure measurements to height estimates expressed in meters. These height measurements are then fused in order to get more accurate estimates than with sonar alone. In the article it is investigated under which conditions fusion is indeed beneficial compared to using only the primary sensor cue.}
\label{fig:overview_SSL}
\end{figure}

Until now, studies on SSL have kept the two sensor cues separated. For example, Stanley did not fuse close-by vision-based road classifications with the laser-based classifications. Fusion in SSL raises several questions. For example, given that the secondary cue is learned on the primary cue's outputs, will its estimates not always be (much) worse? Will the estimates of the secondary cue not be statistically (too) dependent on the primary cue? Given that a ground truth is not available, can the robot determine the uncertainty of the secondary cue reliably enough for successful fusion? The answers to these questions cannot only come from empirical studies on SSL fusion. To answer them in a more generic way, a theoretical investigation is required.



The \textbf{main contribution} of this article lies in a theoretical analysis of the fusion of the primary and secondary cue in SSL. Employing a minimal model of SSL, a theoretical proof is provided that (1) shows that fusion in SSL can indeed lead to better results, and (2) for the given model determines the conditions on the estimation accuracy of the two cues under which fusion is indeed beneficial (Section \ref{section:proof_SSL}). An additional contribution is the verification of the proposed SSL fusion scheme on robotic data (Section \ref{section:case_study}). In particular, SSL is applied to a scenario in which a drone has to estimate its height based on a barometer and a sonar sensor (see Figure \ref{fig:overview_SSL}).

\section{Fusion in self-supervised learning} \label{section:proof_SSL}

\subsection{A minimal model for fusion in self-supervised learning}
\begin{figure}[htp]
\centering
\includegraphics[width=8cm]{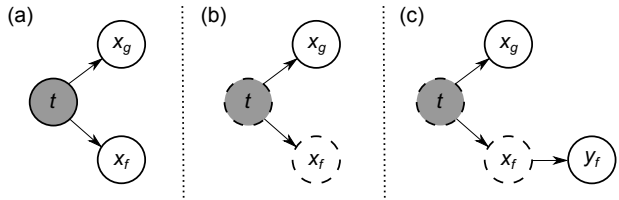}
\caption{Graphical representation of the studied probabilistic self-supervised learning model. (a) The model assumed in the proof. (b) The model from the robot's viewpoint, showing its lack of knowledge. The dashed lines indicate that the robot does not know how $t$ and $x_f$ are distributed. (c) The model showing the knowledge and assumptions on the part of the robot when performing self-supervised learning. In particular, $t$ and $x_f$ are still unknown, but the robot does make an assumption about the distribution of $y_f = f(x_f)$, the secondary cue learned with self-supervised learning.}
\label{fig:probabilistic_model} 
\end{figure}

Figure \ref{fig:probabilistic_model}-(a) shows the graphical probabilistic model used in the proof. The robot has two observations, $x_g$ and $x_f$. From these observations, it will have to infer $t$, which is not observed and therefore shaded in gray. The graphical model shows that $x_g$ and $x_f$ are independent from each other given $t$. Not shown in the figure is the type of distributions from which the variables are drawn. For our minimal model, we will have: $t \sim \mathcal{N}\{0, \sigma_t^2\}$, $x_g \sim \mathcal{N}\{t, \sigma_g^2\}$, and $x_f \sim \mathcal{N}\{t, \sigma_f^2\}$. Figure \ref{fig:probabilistic_model}-(a) is a standard graphical model as can be found in the machine learning literature (e.g., \cite{BISHOP2006}). Typically, it represents the assumptions that the designer and hence the robot has on the structure of the observation task. For the given ground-truth model, the optimal fusion estimate would be $\hat{t} = \argmax_t P(t|x_g, x_f) = \frac{\sigma_g^2 x_f + \sigma_f^2 x_g}{\sigma_g^2 + \sigma_f^2 + \sigma_t^2}$, but this supposes that the robot knows all parameters.

In self-supervised learning, the robot does not have any prior idea of the distribution of the complementary sensory cue. And often, it may not have any idea on the distribution of variable to be estimated either. The reason is that both distributions will likely depend on the unknown environment in which the robot will operate. We represent variables for which the robot does not know the distribution by means of dashed lines in the graphical model. Figure \ref{fig:probabilistic_model}-(b) shows that in our minimal model, the robot does not know the distributions of $t$ and $x_f$. It does know that $x_g \sim \mathcal{N}\{t, \sigma_g^2\}$. 

Self-supervised learning has the robot learn a mapping $f$ from the complementary cue $x_f$ to the trusted cue $x_g$. This leads to a new variable $y_f = f(x_f)$. For this proof, we will have the robot fuse this variable with $x_g$ by making an assumption on the distribution of $y_f$. Hence, in Figure \ref{fig:probabilistic_model}-(c), $y_f$ is shown with a solid line. The dependency of $y_f$ on $x_f$ passes through the function $f$, which in our minimal model is linear with a single parameter $a$: $f(x_f) = a \: x_f$. In our case, the robot will assume that $y_f \sim \mathcal{N}(t, \sigma_{y_f}^2)$. The assumption that $y_f$ is centered on $t$ is, as we will see further below, incorrect, given the ground-truth model. For fusion, the robot needs to know the variance $\sigma_{y_f}^2$. Since it does not know the distributions of $t$ and $x_f$, it will estimate $\sigma_{y_f}^2$ on the basis of the data encountered. The main difficulty here is that the robot evidently does not know what $t$ is for each sample, so the robot will have to use a proxy for the real $\sigma_{y_f}^2$. In our minimal model, the robot will use $\sigma_{y_f \mid x_g}^2$ as a proxy for $\sigma_{y_f}^2$. Finally, please note that the fact that $f$ is learned with the help of $x_g$ does not mean that $y_f$ is conditionally dependent on $x_g$. Obviously, given $x_f$ or $t$, $y_f$ is independent of $x_g$.

\subsection{Proof under which conditions fusion of $y_f$ and $x_g$ leads to better estimates than $x_g$ alone}
Here we will give a closed form solution to the conditions under which a fused estimate $\hat{t}_{\mathrm{fuse}}$ leads to a lower expected squared error than the estimate relying only on $x_g$, denoted by $\hat{t}_g$. We first determine what the estimates are for the two different cases. Please note that in both cases, the robot does not know the distribution of $t$. So, when only using $x_g$, $t$ is estimated by optimizing the likelihood:
\begin{equation}
\hat{t}_g = \argmax_{t} \{ p(x_g \mid t)\} = x_g,
\end{equation}
\noindent where we made use of the fact that the robot knows that $x_g \sim \mathcal{N}\{t, \sigma_g^2\}$. When fusing both cues, $t$ is estimated by optimizing the likelihood of both $y_f$ and $x_g$:
\begin{equation} \label{Eq:fuse_yf}
\hat{t}_{\mathrm{fuse}} = \argmax_t \{ p(y_f, x_g \mid t) \} = \frac{\sigma_g^2 y_f + \sigma_{y_f | x_g}^2 x_g}{\sigma_g^2 + \sigma_{y_f | x_g}^2}.
\end{equation}
In the next subsections, we will use our knowledge of the ground-truth model to determine the associated expected estimation errors. The crux is that this knowledge allows us to predict what function $f$ and what estimate of $\sigma_{y_f \mid x_g}^2$ the robot will converge to given sufficient data. 

\subsubsection{Expected squared error when using $x_g$} 
Here, we determine the expected error when the robot only uses $x_g$: 
\begin{equation}
\mathbb{E}[(\hat{t}_g - t)^2] =  \mathbb{E}[x_g^2 -2 x_g t +t^2],
\end{equation}
\noindent which we can split in the following three parts. First:
\begin{equation}
\mathbb{E}[t^2] = \int_t p(t) t^2 dt = \sigma_t^2,
\end{equation}
\noindent where $\int_t$ is a shorthand for $\int_{t=-\infty}^{\infty}$. Second:
\begin{equation} \label{Eq:E_xg2}
\mathbb{E}[x_g^2] = \int_t \int_{x_g} p(x_g,t) x_g^2 dx_g dt = \int_t p(t) \int_{x_g} p(x_g | t) x_g^2 dx_g dt = \sigma_g^2 + \sigma_t^2,
\end{equation}
\noindent where we made use of $p(x_g | t)$ being Gaussian, $\int_{x_g} p(x_g | t) x_g^2 dx_g = t^2 + \sigma_g^2$. Third:
\begin{equation}
\mathbb{E}[-2 x_g t] = -2 \int_t \int_{x_g} p(x_g, t) x_g  t dx_g dt = -2 \int_t p(t) t \int_{x_g} p(x_g | t) x_g dx_g dt = -2 \int_t p(t) t^2 dt = -2 \sigma_t^2. 
\end{equation}
\noindent The three parts together lead to the expected squared error:
\begin{equation} \label{Eq:fused_error_xg}
\mathbb{E}[(\hat{t}_g - t)^2] = \sigma_t^2 + \sigma_g^2 + \sigma_t^2 - 2 \sigma_t^2 = \sigma_g^2.
\end{equation}

\subsubsection{Expected squared error when fusing $x_g$ and $y_f$}
Here we determine the expected error if the robot fuses $y_f$ with $x_g$. The procedure is as follows. First we express $y_f$ as a function of $x_f$. Then we can retrieve the expression of the fused estimate $\hat{t}_{\mathrm{fuse}}$ and finally calculate the corresponding expected error $\mathbb{E}[(\hat{t}_{\mathrm{fuse}} - t)^2]$.

The function $f$ maps $x_f$ to $x_g$. In the case of our ground-truth model, Figure \ref{fig:probabilistic_model}-(a), what would the parameter $a$ converge to in $f(x_f) = a \: x_f$ if the robot has enough data? Well, we know that $x_g \sim \mathcal{N}\{t, \sigma_g^2\}$, so $\mathbb{E}[x_g | t] = t$. So after many samples, we would expect the function to try and map $x_f$ to $t$. The answer to the question then lies in the calculation of $\mathbb{E}[t | x_f]$. Following \cite{BISHOP2006} (p. 93), given the distributions $p(t) = \mathcal{N}\{0, \sigma_t^2\}$ and $p(x_f | t)=\mathcal{N}\{t, \sigma_f^2\}$:
\begin{equation}
p(t | x_f) = \mathcal{N}\left \{ \frac{\sigma_t^2 x_f}{\sigma_f^2 + \sigma_t^2}, \frac{\sigma_f^2 \sigma_t^2}{\sigma_f^2 + \sigma_t^2} \right \},
\end{equation}  
implying that $\mathbb{E}[t | x_f] = \frac{\sigma_t^2}{\sigma_f^2 + \sigma_t^2} x_f$ and hence:
\begin{equation} \label{Eq:a}
a = \frac{\sigma_t^2}{\sigma_f^2 + \sigma_t^2}.
\end{equation}
We have discussed before that for fusion, the robot will assume that $y_f$ is normally distributed and centered on $t$. This is actually incorrect, as $y_f = a x_f$, with $a \in \langle 0,1 \rangle$. This mapping leads to $y_f \sim \mathcal{N}\{ at, a^2 \sigma_f^2 \}$, which is not centered on $t$. Please remark that this actually makes $y_f$ a better estimate of $t$ than $x_f$ itself, as $y_f$ implicitly takes into account the prior distribution of $t$. 

Next we express the variable $\sigma_{y_f | x_g}^2$ in terms of $\sigma_t$, $\sigma_f$, and $\sigma_g$. From \cite{BISHOP2006} (p. 89) follows:
\begin{equation}
\sigma_{y_f | x_g}^2 = \mathrm{var}(y_f) - \frac{\mathrm{cov}(y_f, x_g)^2}{\mathrm{var}(x_g)}.
\end{equation}
\noindent The variance of $x_g$ is:
\begin{equation}
\mathrm{var}(x_g) = \mathbb{E}[(x_g - \mathbb{E}[x_g])^2],
\end{equation}
\begin{equation} \label{Eq:E_xg}
\mathbb{E}[x_g] = \int_{x_g} p(x_g) x_g dx_g = \int_t p(t) \int_{x_g} p(x_g | t) x_g dx_g d_t = \int_t p(t) t d_t = 0,
\end{equation}
\noindent because $t \sim \mathcal{N}\{0, \sigma_t^2\}$. Hence, $\mathrm{var}(x_g) = \mathbb{E}[x_g^2] = \sigma_g^2 + \sigma_t^2$ (Eq. \ref{Eq:E_xg2}). The variance of $y_f$ is:
\begin{equation}
\mathrm{var}(y_f) = \mathbb{E}[(y_f - \mathbb{E}[y_f])^2],
\end{equation}
\begin{equation} \label{Eq:E_yf}
\mathbb{E}[y_f] = \int_{y_f} p(y_f) y_f dy_f = \int_t p(t) \int_{y_f} p(y_f | t) y_f dy_f dt = \int_t p(t) a t dt = 0,
\end{equation}
\noindent with $a$ from Eq. \ref{Eq:a}. Therefore, the variance of $y_f$ simplifies to:
\begin{equation} \label{Eq:E_yf2}
\mathrm{var}(y_f) = \mathbb{E}[y_f^2] = \int_{y_f} p(y_f) y_f^2 dy_f = \int_t p(t) \int_{y_f} p(y_f | t) y_f^2 dy_f dt
\end{equation}
\begin{equation}
= \int_t p(t) (\sigma_y^2 + a^2 t^2) dt = \sigma_y^2 + \int_t p(t) a^2 t^2 dt = a^2 \sigma_f^2 + a^2 \sigma_t^2 = \frac{\sigma_t^4}{\sigma_f^2 + \sigma_t^2}.
\end{equation}
Finally, the covariance between $y_f$ and $x_g$ is:
\begin{equation} \label{Eq:E_yfxg}
\mathrm{cov}(y_f, x_g) = \mathbb{E}[(y_f - \mathbb{E}[y_f])(x_g - \mathbb{E}[x_g])] = \mathbb{E}[y_f x_g ] = \int_{y_f} \int_{x_g} p(y_f, x_g) y_f x_g dx_g dy_f 
\end{equation}
\begin{equation}
= \int_t p(t) \left( \int_{y_f} p(y_f |t) y_f dy_f \right) \left( \int_{x_g} p(x_g | t) x_g dx_g \right) dt = \int_t p(t) \left( a t \right) \left( t \right) dt = a \sigma_t^2 = \frac{\sigma_t^4}{\sigma_f^2 + \sigma_t^2}.
\end{equation}
Putting all of this together, we have:
\begin{equation} \label{Eq:sigma_yf_estimated}
\sigma_{y_f | x_g}^2 = \frac{\sigma_t^4}{\sigma_f^2 + \sigma_t^2} - \frac{(\sigma_t^8)}{(\sigma_t^2 + \sigma_f^2)^2(\sigma_t^2 + \sigma_g^2)}.
\end{equation}

The formula for the robot's expected square error is:
\begin{equation} \label{Eq:exp_fuse_yf}
\mathbb{E}[(\hat{t}_{\mathrm{fuse}} - t)^2] = \mathbb{E}[\hat{t}_{\mathrm{fuse}}^2 -2 \hat{t}_{\mathrm{fuse}} t + t^2],
\end{equation}
\noindent where we will write $\hat{t}_{\mathrm{fuse}}$ as $\alpha y_f + \beta x_g$, with $\alpha$ and $\beta$ the factors from Eq. \ref{Eq:fuse_yf}. The expectation in Eq. \ref{Eq:exp_fuse_yf} can be split up in three parts. First:
\begin{equation}
\mathbb{E}[\hat{t}_{\mathrm{fuse}}^2] = \mathbb{E}[\alpha^2 y_f^2 + 2 \alpha \beta y_f x_g + \beta^2 x_g^2] = \alpha^2 (a^2 \sigma_y^2 + a^2 \sigma_t^2) + 2 \alpha \beta (a \sigma_t^2) + \beta^2 (\sigma_g^2 + \sigma_t^2),
\end{equation}
\noindent where we made use of Eqs. \ref{Eq:E_yf2}, \ref{Eq:E_yfxg}, and \ref{Eq:E_xg2}. Second:
\begin{equation}
\mathbb{E}[-2 \hat{t}_{\mathrm{fuse}} t] = -2 \int_t \int_{y_f} \int_{x_g} p(y_f, x_g, t) (\alpha y_f + \beta x_g) t dx_g dy_f dt
\end{equation}
\begin{equation}
= -2 \int_t p(t) t \left( \int_{y_f} p(y_f | t) \alpha y_f dy_f + \int_{x_g} p(x_g | t) \beta x_g dx_g \right) dt = -2 \int_t p(t) t \left( \alpha a t + \beta t \right) dt = -2 (\alpha a + \beta) \sigma_t^2
\end{equation} 
And third: $\mathbb{E}[t^2] = \sigma_t^2$, since $t \sim \mathcal{N}\{0, \sigma_t^2\}$.

Putting these formulas together into Eq \ref{Eq:exp_fuse_yf} and simplifying, gives:
\begin{equation} \label{Eq:fused_error_xg_yf}
\mathbb{E}[(\hat{t}_{\mathrm{fuse}} - t)^2] = \frac{\sigma_g^2 \sigma_t^2 (\sigma_g^2 \sigma_f^4 + \sigma_g^2 \sigma_f^2 \sigma_t^2 + \sigma_t^6 + \sigma_g^2 \sigma_t^4)}{(\sigma_t^4 + \sigma_g^2 \sigma_t^2 + \sigma_f^2 \sigma_g^2)^2}.
\end{equation}

\subsubsection{When fusion is better than just using $x_g$:}
In order to prove that a robot employing self-supervised learning can obtain better estimates of $t$ than when using only $x_g$, we only need to show that there are conditions in which the expected error of Eq. \ref{Eq:fused_error_xg_yf} is smaller than that of Eq \ref{Eq:fused_error_xg}. Given $\sigma_t^2,\sigma_g^2,\sigma_f^2 > 0$, the expected fused error is smaller if:
\begin{equation} \label{Eq:low_sigma_t}
\sigma_t^2 \leq \sigma_g^2,
\end{equation}
or else ($\sigma_t^2 > \sigma_g^2$) if:
\begin{equation} \label{Eq:low_sigma_f}
\sigma_f^2 < -\frac{1}{2} \frac{2\sigma_g^4\sigma_t^2+3\sigma_t^4\sigma_g^2+\sigma_t^6+\sqrt{17\sigma_g^4\sigma_t^8+18\sigma_g^2\sigma_t^{10}+\sigma_t^{12}}}{\sigma_g^4-\sigma_t^4}
\end{equation}

Intuitively, these conditions correspond to (i) $t$ having a strong prior (Eq. \ref{Eq:low_sigma_t}) or (ii) $x_f$ being sufficiently informative on $t$ (Eq. \ref{Eq:low_sigma_f}). The first case of the strong prior may not be easy to understand. It helps to think of the fact that the learned secondary cue $y_f$ takes the prior into account, while $x_g$ does not (as we assume that the robot does not know anything about the prior distribution of $t$). Therefore, fusion with $y_f$ is more advantageous if the prior is stronger. 

To summarize, while the robot wrongly assumes $y_f$ to be centered on $t$ and does not know the real $\sigma_y^2$, it can outperform just using $x_g$ under the conditions in equations \ref{Eq:low_sigma_t} and \ref{Eq:low_sigma_f}. 

\subsubsection{Computational Verification}
The theoretical findings above were verified with computational experiments, in which a data set $\mathcal{D}= \{ (t_1, x_{g1}, x_{f1}), \ldots, (t_N, x_{gN}, x_{fN}) \}$ was generated according to the ground truth model from Figure \ref{fig:probabilistic_model}-(a). Then, the program first learned the parameter $a$ of function $f$ with least-squares regression. Subsequently, it estimated $\sigma_{y_f | x_g}^2$ by determining the variance of $y_f$ when $x_g$ is in the interval of $[-0.05, 0.05]$. Finally, it fused the observations $x_g$ and $y_f$ according to Eq. \ref{Eq:fuse_yf}. 

The error is compared to that of using $x_g$ alone. Given a large enough $N$ the results converge to the values predicted in the theoretical analysis. For instance, with $\sigma_t^2=6.25$, $\sigma_g^2 = 1$, $\sigma_f^2 = 1$, and $N=10,000$, we get a fused squared error of $0.47$ (theoretical prediction $0.49$). With $x_g$ alone the error is $0.98$ (theoretical prediction $1.00$). The theoretical threshold on fusion not being useful anymore is $\sigma_f^2 > 11.57$. Please note that this is a rather benign condition, as $\sigma_f^2$ can be more than 11 times as large as $\sigma_g^2$ in this case. Table \ref{table:computational_results} shows results for four different instances. The bottom case illustrates that fusion helps if the prior is strong enough, even if $\sigma_f^2$ is high. The MATLAB code is part of the supplementary material.


\begin{table}
\caption{Computational results verifying the theoretical proof.} \label{table:computational_results} 
\centering
\begin{tabular}{ | c c c | c | c |} 
\hline
$\sigma_t^2$ & $\sigma_g^2$ & $\sigma_f^2$ & Error primary (theory) & Fusion error (theory)\\
\hline
6.25 & 1 & 1 & $0.98$ ($1.00$) & $0.47$ ($0.49$) \\
6.25 & 1 & 16 & $1.02$ ($1.00$) & $1.11$ ($1.15$) \\
0.25 & 1 & 1 & $1.00$ ($1.00$) & $0.18$ ($0.18$) \\
0.25 & 1 & 100 & $1.00$ ($1.00$) & $0.25$ ($0.25$) \\
\hline
\end{tabular}
\end{table}










\section{Case study: Height estimation with a barometer and sonar.} \label{section:case_study}
In this section, we apply SSL fusion to a case study, in which a flying robot uses a barometer and sonar to estimate the height. As human designers we know how these sensors relate to the height, but in the case study we will assume that the robot only knows how to relate one of the sensors to the height (assumed to be $x_g$), and will regard the other sensor as the ``unknown'' $x_f$. The main goal of the case study is to see if fusion in an SSL setup can be beneficial in a real-world case, which may not comply with the assumptions of the theoretical analysis. A scenario with two scalar measurements was chosen in order to allow a direct comparison with the theoretical model.


\subsection{Experimental setup}
A Parrot AR drone $2.0$ is used for gathering the experimental data. The drone is flown inside of a motion tracking arena. It uses the open source autopilot Paparazzi (\cite{HATTENBERGER2014}) to log the relevant sensor data, consisting of the pressure, the sonar readings, and the height provided by the motion tracking system. The height from the Optitrack motion tracking system is considered the most reliable of the three sensors and hence is used in this case as the `ground-truth' value ($t$). 



The sonar measurements can be directly used as primary cue. If the pressure measurements are used as primary cue, they are mapped to a height estimate $h'_P$ in meters with the following formula:
\begin{equation} \label{Eq:pressure_height}
h'_P = \frac{R T_s}{M g} \mathrm{log}(\frac{P_s}{P}),
\end{equation}
\noindent where $R = 8.31446$ is the gas constant, $T_s = 288.15$ is the sea level temperature, $M = 0.0289644$ is the molar mass of the Earth's air, $g = 9.80665$ the gravity, $P_s = 101325.0$ the sea level pressure, and $P$ the measured pressure. After this conversion to height, there is still an offset and scaling factor due to the fact that the drone has not been flying in the exact circumstances represented by the constants (at sea level for instance). Typically, this offset is taken into account by calibrating the pressure measurement at take-off. Here, $h'_P$ is mapped with a linear function to the Optitrack height (on the training set). The resulting heights $h_P$ are then used as the target values in the self-supervised learning, i.e., as the $x_g$ in the theoretical analysis.

\begin{figure}[htp]
\centering
\includegraphics[width=6.5cm]{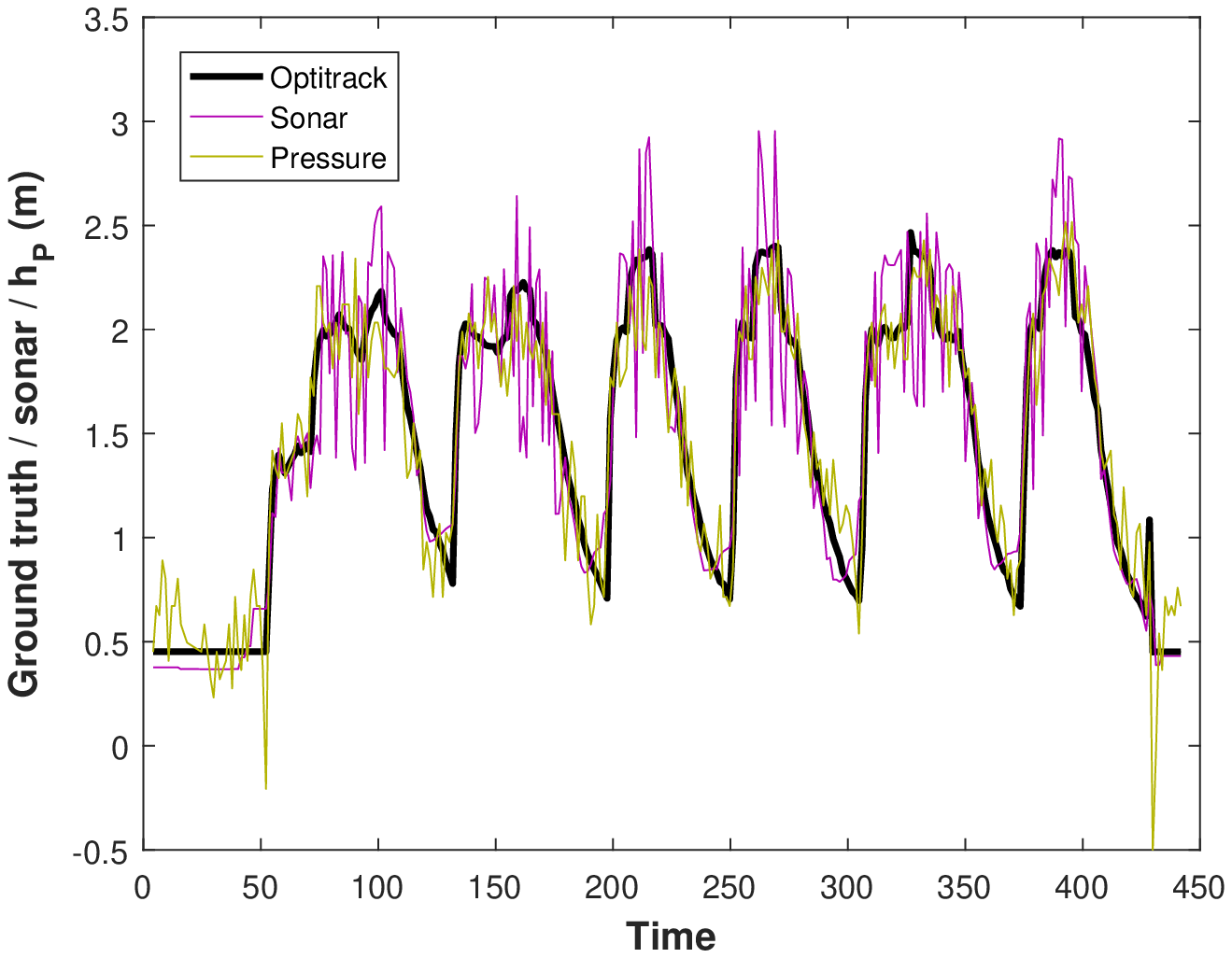} \includegraphics[width=6.5cm]{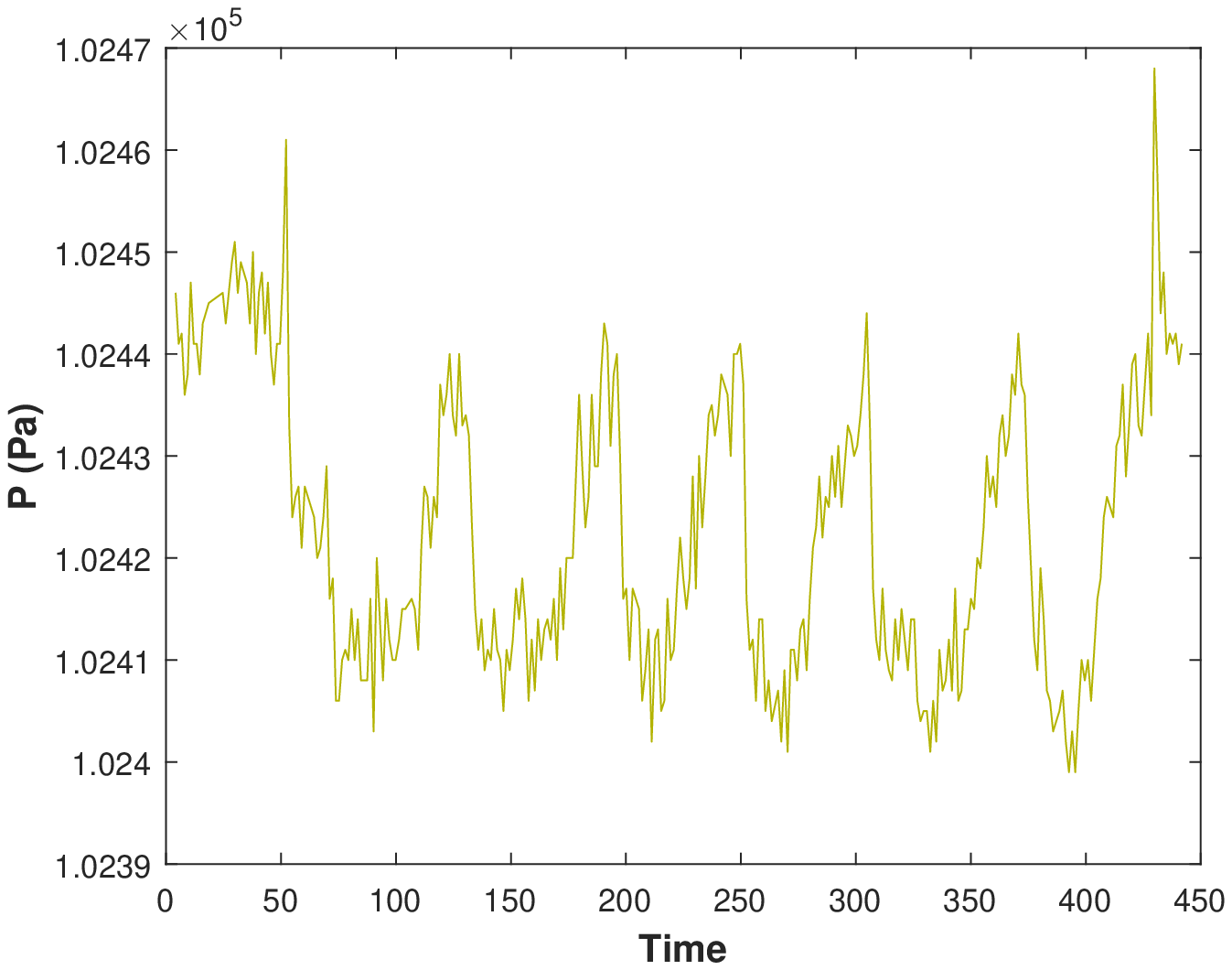}
\caption{Drone data. Left plot: height estimations from the Optitrack motion tracking system, sonar, and barometer. Right plot: distribution of raw pressure measurements from the barometer.}
\label{fig:data}
\end{figure}

Figure \ref{fig:data} gives insight into the data. The left plot shows the Optitrack ground truth height (thick black line), the sonar (purple line), and the corrected barometer measurements when used as primary cue, $h_P$ (dark yellow line). The right plot shows the untransformed pressure measurements $P$. These `raw' measurement values are used when the barometer represents the secondary cue. The magnitude of these measurements already shows that the distribution of pressure measurements is not centered at $t$, as assumed in the theoretical analysis.




The secondary cue is mapped to the primary cue with a machine learning method, which performs regression on the training set. Here we use a $k$ nearest neighbor approach, with $k = 3$, so that possible non-linear relations can be captured (for instance from the raw pressure measurements to the sonar height). Furthermore, in the experiment the standard deviation of $x_g$ is assumed to be known - and is here determined with the help of the ground truth $t$ from Optitrack on the training set. The conditional standard deviation $\sigma_{y_f|x_g}$ is determined on the validation set, only using the variables observed by the robot, $y_f$ (the secondary cue obtained with regression function) and $x_g$ (the primary cue). For each condition, we perform $N = 100$ experiments on the data. In each experiment, $80\%$ of the data set is used for training, $10\%$ of the set is used for validation (determining the conditional standard deviation $\sigma_{y_f|x_g}$), and $10\%$ is used for testing. For each experiment, we determine the mean absolute error of the primary cue $x_g$, that of $y_f$, and their fusion. 

\subsection{Experimental results}
There are two different experimental conditions: (i) the sonar is the primary cue and the barometer the secondary cue, and (ii) vice versa. Table \ref{table:case_study_results} shows the main results from the experiments. In both conditions, the SSL fusion consistently gives (slightly) better results than just using the primary cue. 


\begin{table}
\caption{Main results of the SSL fusion experiment.} \label{table:case_study_results}
\centering
\begin{tabular}{ | l | c c c | c |} 
\hline
& & Mean Absolute Error (m) & & \\
Primary & Primary ($x_g$) & Secondary ($x_f$) & Fusion & Successful fusion\\
\hline
Sonar & $0.22$ & $0.20$ & $0.17$ & $100\%$\\
Barometer & $0.15$ & $0.19$ & $0.14$ & $98\%$\\
\hline
\end{tabular}
\end{table}

\begin{figure*}[tb!]
\centering
\includegraphics[width=4.5cm]{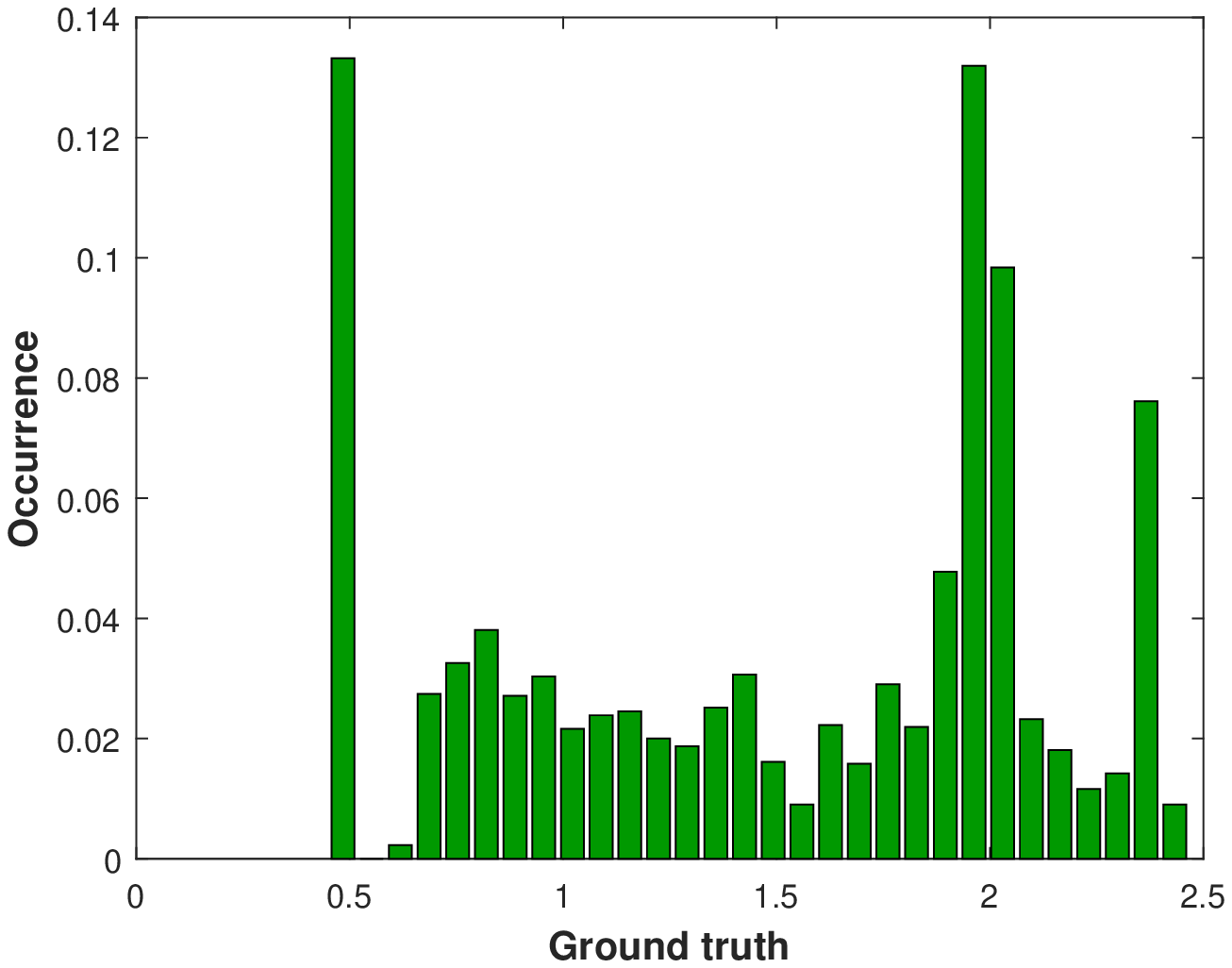} \includegraphics[width=4.5cm]{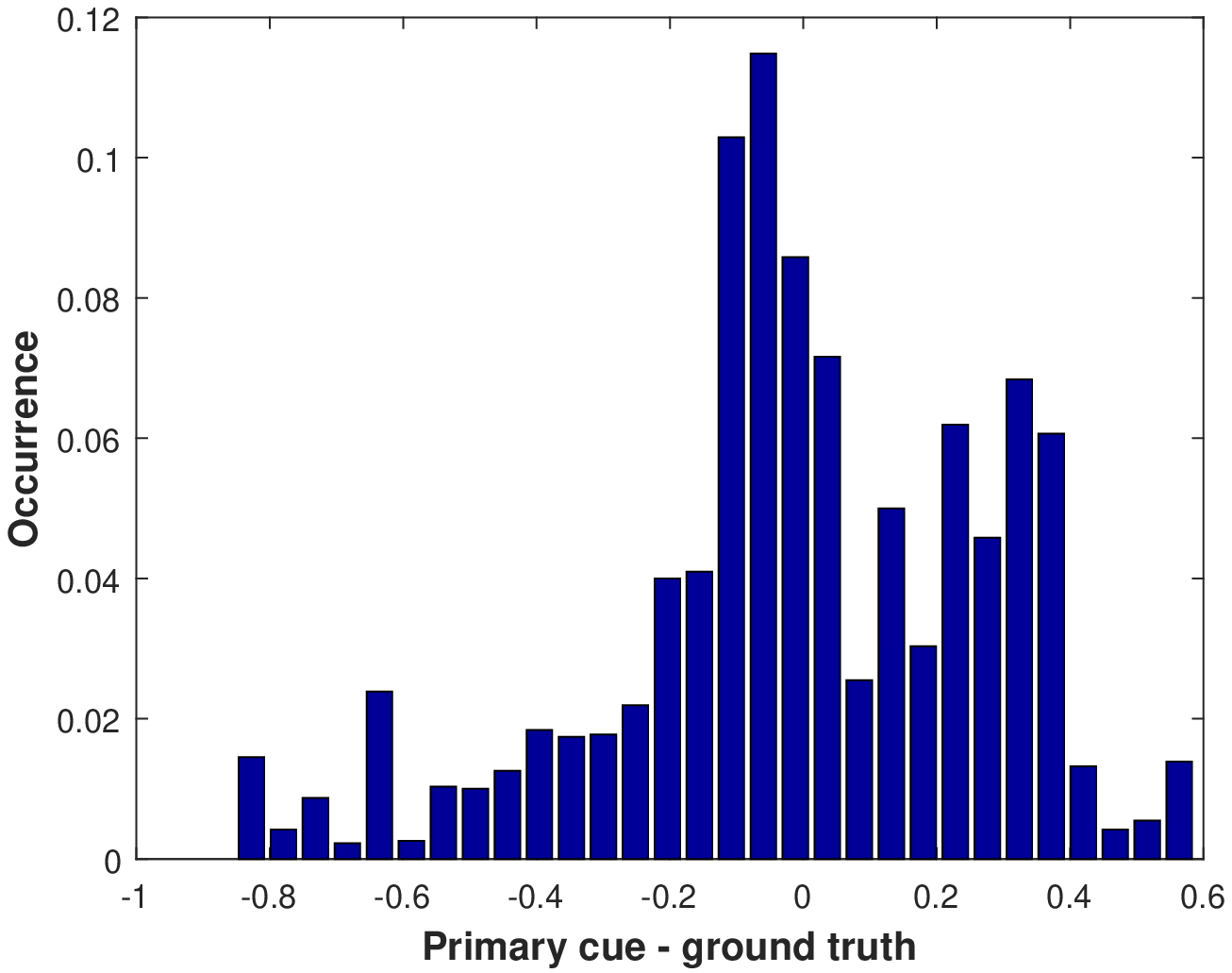} \includegraphics[width=4.5cm]{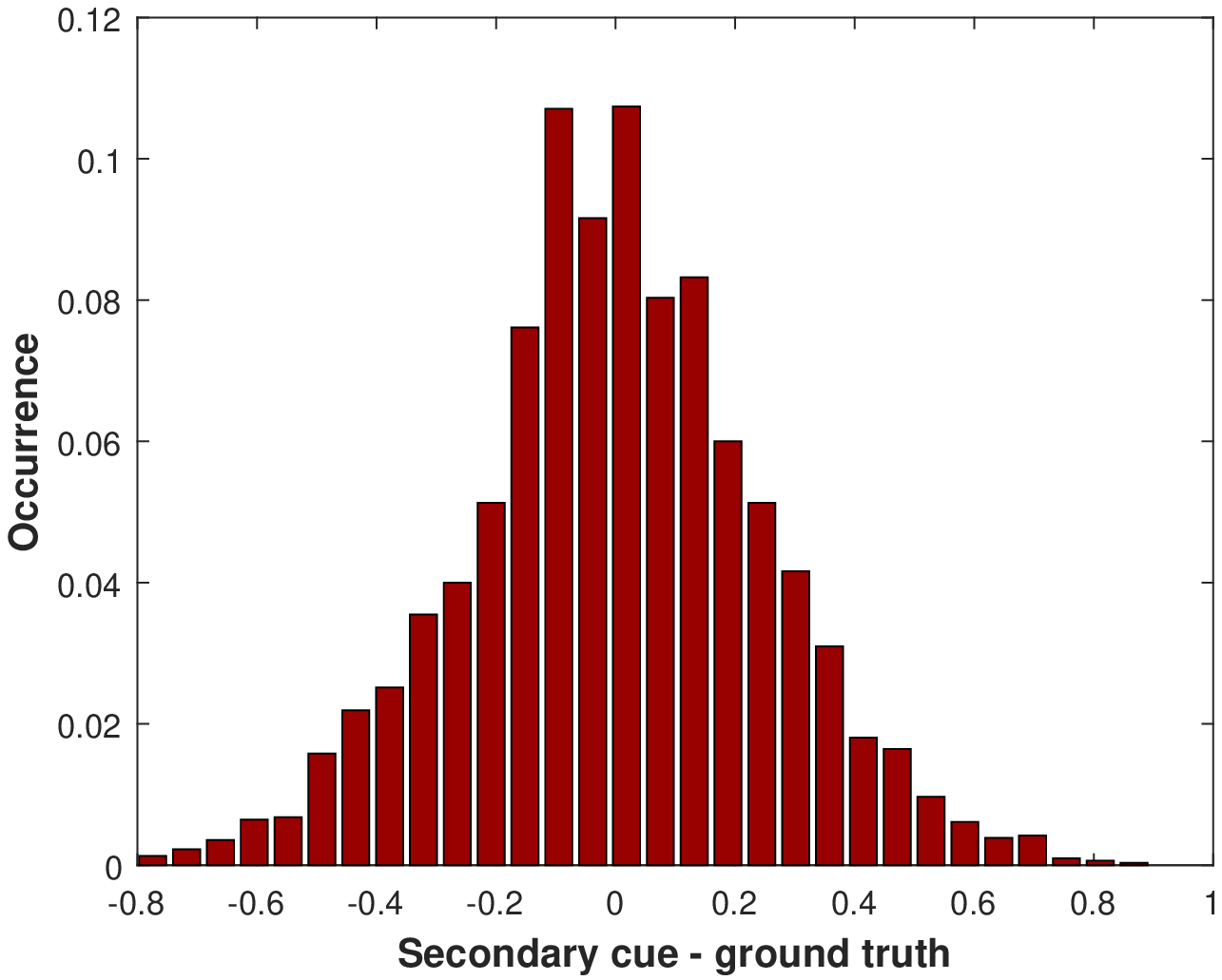}
\includegraphics[width=4.5cm]{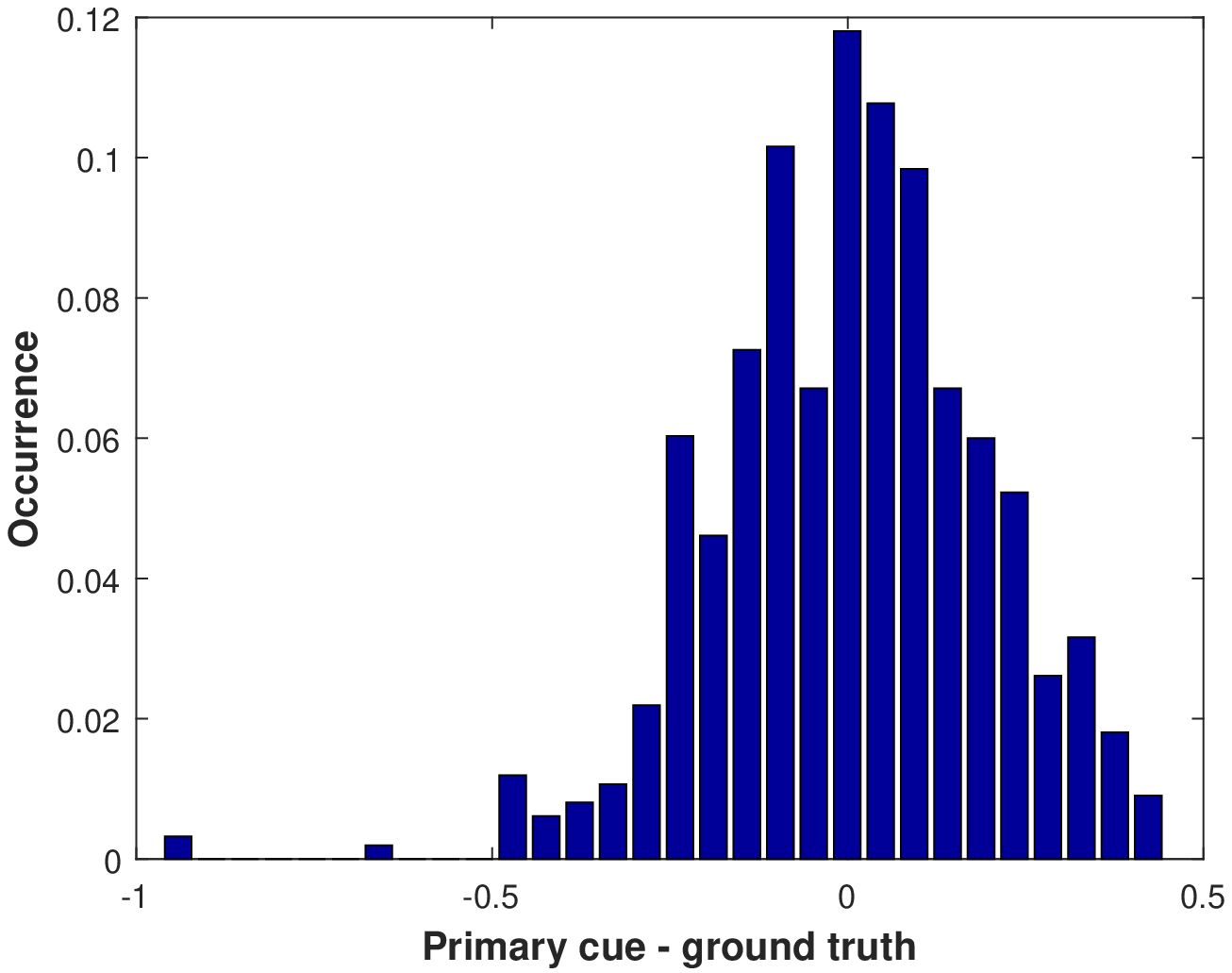} \includegraphics[width=4.5cm]{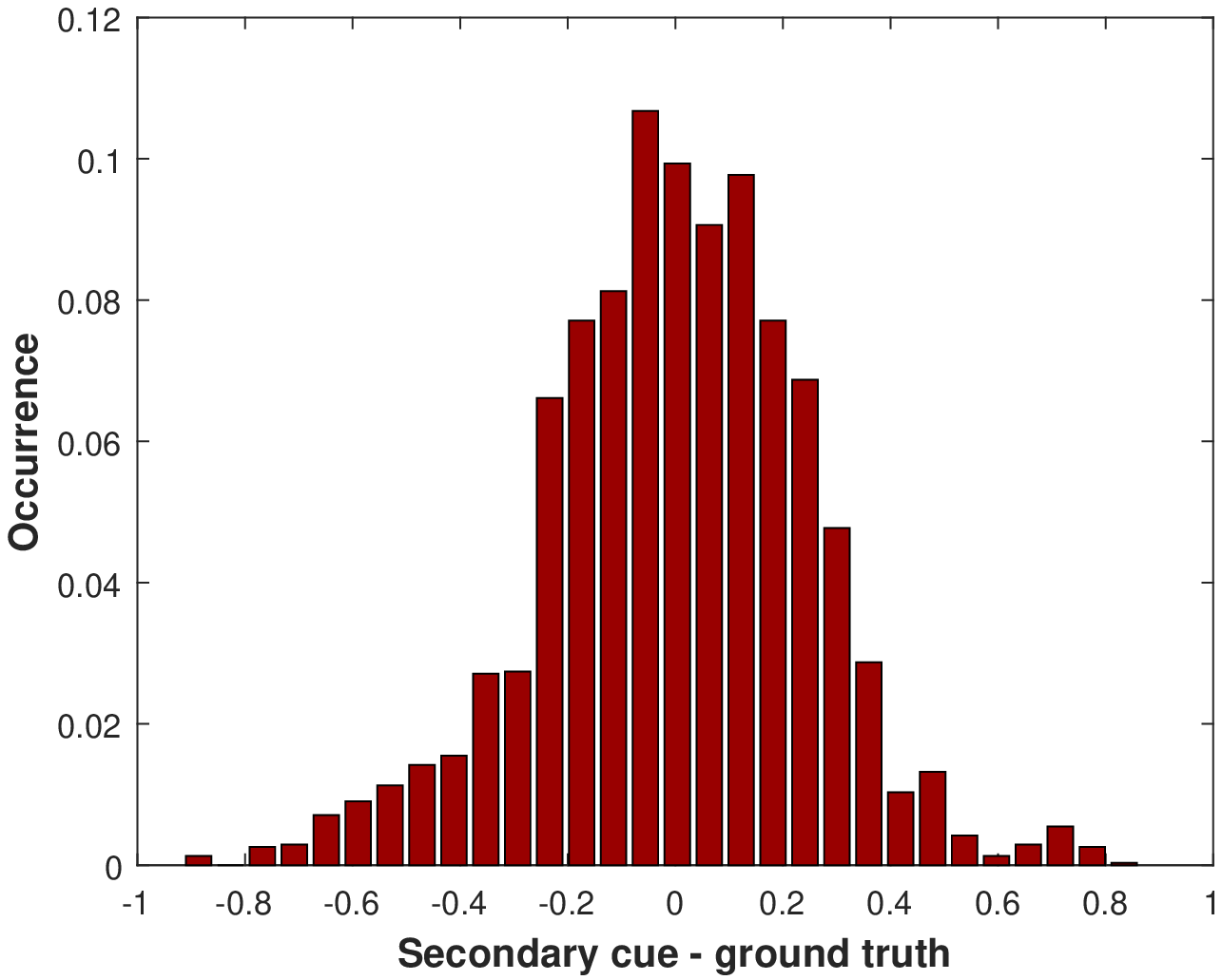}
\caption{Distributions of the relevant variables. \textbf{Top row:} Sonar is the primary cue. From left to right: Distribution of the Optitrack groundtruth height $t$, distribution of the error of the sonar primary sensor cue $x_g - t$, and distribution of the error of the pressure-based height learned with SSL, $y_f - t$. \textbf{Bottom row:} Pressure is the primary cue. Left: Distribution of $x_g - t$. Right: Distribution of $y_f - t$.}
\label{fig:distributions}
\end{figure*}

Let us analyze the case where the sonar is the primary cue in order to see how well the distributions of the involved variables correspond to the assumptions in our minimal model. Figure \ref{fig:distributions} (top row) shows the distributions of the Optitrack height ($t$), the error of the sonar height ($x_g - t$), and the error of the pressure-based height estimate learned with SSL ($y_f - t$). The corresponding means and standard deviations are: $\mu_t = 1.48$, $\sigma_t = 0.64$, $\mu_g = -0.001$, $\sigma_g = 0.29$, $\mu_{y_f} = 0.01$, and $\sigma_{y_f} = 0.25$. These numbers show that both the primary and secondary cue are centered on $t$, and that the accuracy of the secondary cue is actually better than that of the primary cue ($\sigma_y < \sigma_g$). We compared each distribution against a normal distribution that has the same mean and standard deviation. The Chi-square values are $0.78$, $0.35$, and $0.03$ for $t$, $x_g-t$ and $y_f-t$, respectively, confirming that the secondary cue indeed resembles its corresponding normal distribution most. However, a randomized statistical test (\cite{COHEN1995}) shows that even the histogram of $y_f-t$ is unlikely to come from the corresponding normal distribution (with a $p$-value of $8^{-4}$).


An analysis of pressure as the primary cue paints a similar picture. Two things are interesting to observe though. The first observation is that in this condition, the secondary cue is less accurate than the primary cue; $\sigma_g = 0.19$ and $\sigma_y = 0.25$. The second observation follows from the bottom row of Figure \ref{fig:distributions}, which shows the distributions when pressure is the primary cue. The right plot shows the distribution of sonar as a secondary cue. The distribution, $y_f$ in this condition, seems much more normally distributed than when sonar is the primary cue (top row). Indeed, the Chi-square value is $0.002$ for both the primary and secondary cue in this condition (still with a low $p$-value of $10^{-5}$).

To summarize the findings of the analysis, the variables in the real-world experiment deviate from the model's assumptions of how they are distributed. Despite this, fusion still leads to better results. It may be though that the threshold value differs from the theoretical one. This is akin to using a Kalman filter when the involved distributions are not normal; The filter will most of the time still give a reasonable result, but estimation optimality is no longer guaranteed. Interestingly, this case study shows that the threshold expressed in Eq. \ref{Eq:low_sigma_f} often cannot be validated. It would for instance not be very useful to look at $\sigma_f$ when pressure is the secondary cue, as it has wildly different values from $t$. It may be better to express the threshold in Eq. \ref{Eq:low_sigma_f} in terms of $\sigma_{y_f}$. This can be done by using the relation $\sigma_{y_f}^2 = a^2 \sigma_f^2$, with $a$ defined in Eq. \ref{Eq:a}. If the terms on the right-hand side of Eq. \ref{Eq:low_sigma_f} are represented by the variable $C$, this leads to the threshold: $\sigma_{y_f}^2 < a^2 C$.

\section{Conclusions} \label{section:conclusions}
In this article, a theoretical analysis was performed under which conditions it is favorable to fuse the primary and secondary cue in self-supervised learning. This analysis shows that fusion of the cues with the robot's knowledge is favorable when (i) the prior on the target value is strong, or (ii) the secondary cue is sufficiently accurate. When the assumptions of the analysis are valid, the conditions for the usefulness of fusion are rather benign. In the studied model, the standard deviation of the secondary cue can be more than ten times that of the primary cue, while still giving better fusion results. Although the employed model is rather minimal, the result that fusion can lead to better estimates extends to more complex cases, as is confirmed by the real-world case study. However, violations of the assumptions will likely change the threshold on the secondary cue's accuracy.

Given that normal distributions approximate quite well various real-world phenomena, the theoretical analysis may be applicable to a wide range of cases. Still, the generalization of the main finding - that SSL fusion can give better results than the primary cue alone - to more complex cases should be further investigated. To this end, future work could employ the current proof as a template. Moreover, it would be interesting to apply SSL fusion to a more complex, relevant case study than the one studied here. For instance, it would be highly interesting if SSL fusion could improve the performance of complex senses such as robotic vision. 



\bibliographystyle{plain}
\bibliography{bibliogdc}

\end{document}